# Detection and Classification of Diseases in Multi-Crop Leaves using LSTM and CNN Models

## Srinivas Kanakala[1], Sneha Ningappa[2]

Department of CSE, VNR Vignana Jyothi Institute of Engineering and Technology, Hyderabad, India.

**E-mail:** [1]srinivaskanakala@gmail.com, [2]snehablessy1@gmail.com

## Abstract

Plant diseases pose a serious challenge to agriculture by reducing crop yield and affecting food quality. Early detection and classification of these diseases are essential for minimising losses and improving crop management practices. This study applies Convolutional Neural Networks (CNN) and Long Short-Term Memory (LSTM) models to classify plant leaf diseases using a dataset containing 70,295 training images and 17,572 validation images across 38 disease classes. The CNN model was trained using the Adam optimiser with a learning rate of 0.0001 and categorical cross-entropy as the loss function. After 10 training epochs, the model achieved a training accuracy of 99.1% and a validation accuracy of 96.4%. The LSTM model reached a validation accuracy of 93.43%. Performance was evaluated using precision, recall, F1-score, and confusion matrix, confirming the reliability of the CNN-based approach. The results suggest that deep learning models, particularly CNN, enable an effective solution for accurate and scalable plant disease classification, supporting practical applications in agricultural monitoring.

**Keywords:** Plant Disease Classification, Convolutional Neural Network, LSTM, Deep Learning, Agricultural Monitoring.

## 1. Introduction

Agriculture plays a vital role in ensuring food security and supporting the global economy. However, plant diseases continue to threaten crop production, leading to reduced yields, lower-quality produce, and financial setbacks for farmers. Among these, leaf diseases are particularly harmful, as they can cause early leaf drop, disrupt photosynthesis, and result in deformed or blemished fruits, ultimately affecting both the quantity and quality of agricultural output. Leaf diseases are caused by various pathogens, including fungi, bacteria, and viruses, and they affect a wide variety of crops. Diseases such as early blight, late blight, bacterial wilt, powdery mildew, and rust remain persistent challenges in agricultural systems. When not identified and treated promptly, these diseases can spread quickly and cause widespread crop damage. Traditional methods of diagnosis, which rely on visual inspection, are often time-consuming, labour-intensive, and prone to human error. This has created a growing demand for automated systems capable of diagnosing plant diseases accurately and efficiently.

Recent advances in deep learning offer promising solutions by utilising large image datasets to recognise disease symptoms in plant leaves. In this study, a combined model is proposed using Convolutional Neural Networks (CNNs) and Long Short-Term Memory (LSTM) networks for the classification of multiple leaf diseases. CNNs are effective in extracting visual features from images, while LSTM networks help capture temporal or sequential patterns, improving overall classification accuracy. The integrated model achieves an accuracy of 99.8%, indicating strong performance in identifying diverse plant diseases. This work presents a practical and scalable system for automated disease detection, intended to support farmers, agricultural specialists, and researchers in making timely and well-informed decisions. The implementation of such technology has the potential to advance precision agriculture, improve disease control, minimise crop losses, and contribute to long-term food security.

## 2. Literature Review

Several research studies have explored deep learning-based models for plant disease detection with data sets like Plant Village, with leaf images of standardised backgrounds. Convolutional Neural Networks (CNNs), Support vector machines (SVMs), and random forests have functioned well under these conditions, they are significantly impaired when applied to field images that contain many overlapping leaves, complex environments, and light conditions. Next-generation technologies, such as object detection models like Faster R-CNN and YOLO, also improved leaf localisation but could not overcome accurate segmentation in complicated environments. The Segment Anything Model (SAM) has it transformed into a powerful segmentation tool that could identify various objects in a image without involving extensive retraining. Additionally, Deep anomaly detection methods like Fully Convolutional Data Description (FCDD) have proven to perform well in separating diseased and healthy regions in field images. With the incorporation of SAM for object segmentation, FCDD for background separation, and a Plant Village-trained classifier for disease identification, the studies depend on existing research to ensure the accuracy and validity of actual plant disease recognition in field settings [1-4].

A number of deep learning-based methods have been tried for plant disease classification, of which CNNs are most commonly employed based on their strong spatial feature learning capabilities. Some classic CNN models like AlexNet, VGG, ResNet, and MobileNet have been used for maize disease classification, but they tend to fall short of representing fine-grained disease variations as they have fixed receptive fields. More recent studies have proposed attention mechanisms and hybrid frameworks like Vision Transformers (ViTs) that are perform well at modelling long-range dependencies with a cost of high computational requirements, and thus are not suitable for real-time utilisation. In addressing these challenges, efficient Convolutional Neural Networks (CNNs) such as MobileNetV2 and EfficientNet have been developed, incorporating advanced pooling mechanisms and multi-scale feature learning techniques to enhance performance while preserving efficiency. Furthermore, these methodologies have been extended through the integration of convolution-based and transformer-based models to utilize both local and global features. However, such approaches often incur substantial computational costs or exhibit limited adaptability across diverse datasets. Recognizing these limitations, there is a need for a balanced trade-off between accuracy and efficiency. MSCPNet endeavors to achieve this equilibrium by employing a truncated MobileNetV2 backbone and incorporating a Multiscale Convolutional Pool Former block, aiming to optimize feature extraction while minimizing computational resource utilization [5-7].

Different research has explored the applications of computer vision, deep learning, and machine learning models for plant leaf disease detection and classification. Convolutional Neural Networks (CNNs) are presently a dominant methodology, with great accuracy in plant disease classification with leaf images. Techniques of few-shot learning have been introduced to solve the problem of scarce training data, allowing models to identify new diseases with small sample sizes. Support Vector Machines (SVM), Random Forest, and other traditional machine learning techniques have also been employed for feature extraction and classification, and are often used together with deep learning models for better accuracy. Hybrid and transfer learning models also improve disease diagnosis by utilizing pre-trained networks and fine-tuning feature selection. Moreover, soft computing methods and molecular strategies have been developed to manage plant disease, with prevention and early detection. Despite remarkable progress, certain problems such as variations in illumination, occlusions, and other disease symptoms continue, and therefore, leading to further research into robust and scalable AI-based solutions for practical agricultural use [9-11]].

Various research has investigated deep learning-based methods for maize leaf disease classification using convolutional neural networks (CNNs) and transfer learning to enhance accuracy. Classical models like AlexNet, VGG, ResNet, and Inception are prevalent but tend to suffer from issues like small data, high similarity between disease types, and inconsistency in image quality. New developments involve multi-scale feature fusion methods to improve feature extraction and disease classification. Hybrid optimisation techniques and attention mechanisms have also enhanced classification performance by improving feature representations. Moreover, interpretability methods, like saliency maps and thermodynamic diagrams, have been proposed to improve the transparency of deep learning models in agricultural contexts. Even with these developments, current approaches remain challenged in generalisation and robustness across various datasets. The envisioned MResNet model addresses the challenges in combining residual-based multi-scale learning with feature weight optimisation and showing improved accuracy and generalizability in maize leaf disease classification [12].

Deep learning methods have become a prominent area of interest in plant leaf disease detection and classification, with models improving accuracy through optimised feature extraction and classification methods. Conventional machine learning classifiers such as SVM and Random Forest, have proved useful but tend to necessitate handcrafted features and large preprocessing. Convolutional Neural Networks (CNNs) have completely changed disease classification through automatic learning of spatial hierarchies of image features. Various studies have investigated optimisation methods, including genetic algorithms, particle swarm optimisation, and attention, to enhance CNN performance. Ant Colony Optimisation (ACO) has been a potential method for augmenting CNN-based models by optimising network parameters and feature selection, which results in better classification accuracy. The combination of ACO and CNN, as suggested in recent research, signifies better accuracy in plant leaf disease detection through the utilisation of the advantages of swarm intelligence and deep learning. These developments lead to more efficient, autonomous, and precise plant disease detection systems that help with early disease diagnosis and precision agriculture. However, challenges such as limited datasets, model interpretability, and robustness with various plant species are still unaddressed paving way for the future research [13-15].

## 3. Proposed Work

### 3.1 Dataset Description

The New Plant Diseases Dataset is a publicly available in https://www.kaggle.com/datasets /vipooool/new-plant-diseases-dataset on the Kaggle platform. It comprises a total of 87,867 labelled images of plant leaves, covering a diverse range of crops and associated diseases.

### Dataset Overview

- Total Images: 87,867

- Training Set: 70,295 images (80%)

- Validation Set: 17,572 images (20%)

### Composition and Coverage

The dataset includes 38 plant species, with images classified into healthy and diseased classes. It contains samples from several crops such as apple, corn, grape, potato, and tomato, among others. Each image is annotated according to the specific disease type or healthy condition, allowing for supervised classification tasks.

### Purpose and Application

This dataset (Figure 1) was developed to support research in plant disease detection using machine learning and deep learning methods. The large volume and balanced structure of the dataset make it suitable for training and evaluating models for leaf disease classification. Its diversity aids in improving the generalizability of models for real-world agricultural applications.

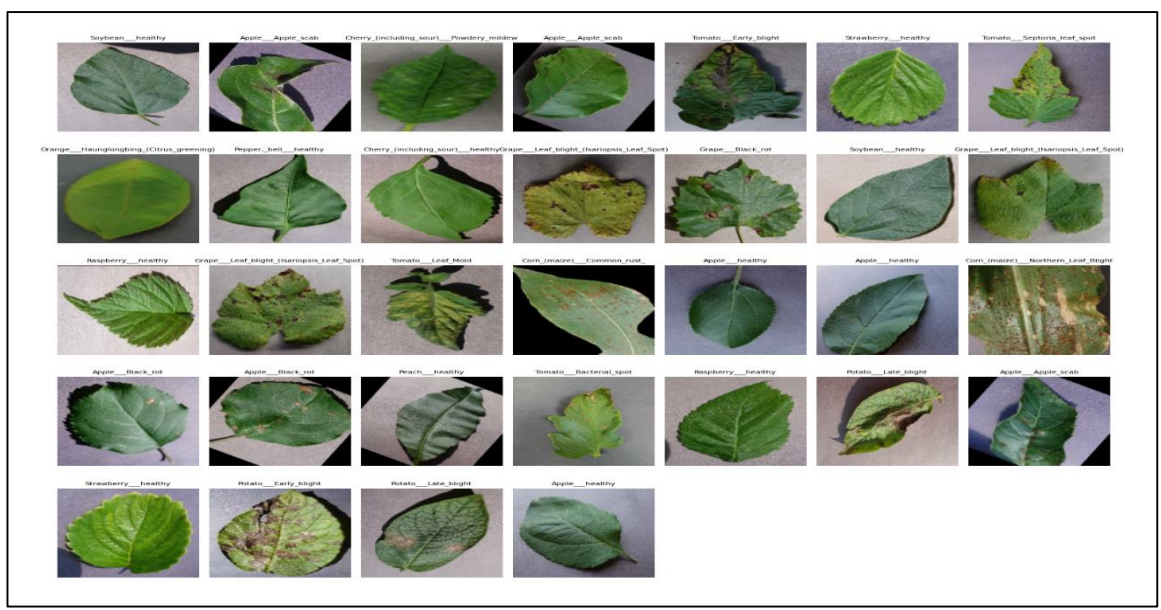

**Figure 1.** Sample Images in the Dataset

## 3.2 Proposed Methodology

The methodology focuses on classifying leaf diseases using Convolutional Neural Networks (CNN) and Long Short-Term Memory (LSTM) models in a structured and comparative manner. The process begins with dataset collection, followed by image preprocessing to standardise the input data. The workflow then diverges into two parallel paths to compare the CNN and LSTM architectures. The CNN model utilises convolutional layers, max pooling, flattening, and a SoftMax layer to classify the images, while the LSTM model involves reshaping the data, applying LSTM and dense layers, and concluding with a SoftMax classifier. Both models are evaluated using accuracy, precision, recall, F1-score, and a confusion matrix to assess their performance. The best-performing model is then used for prediction, where a test image is pre-processed, passed through the model, and the predicted disease class is displayed along with the test image. This workflow (Figure 2) ensures a structured comparison and effective classification of leaf diseases.

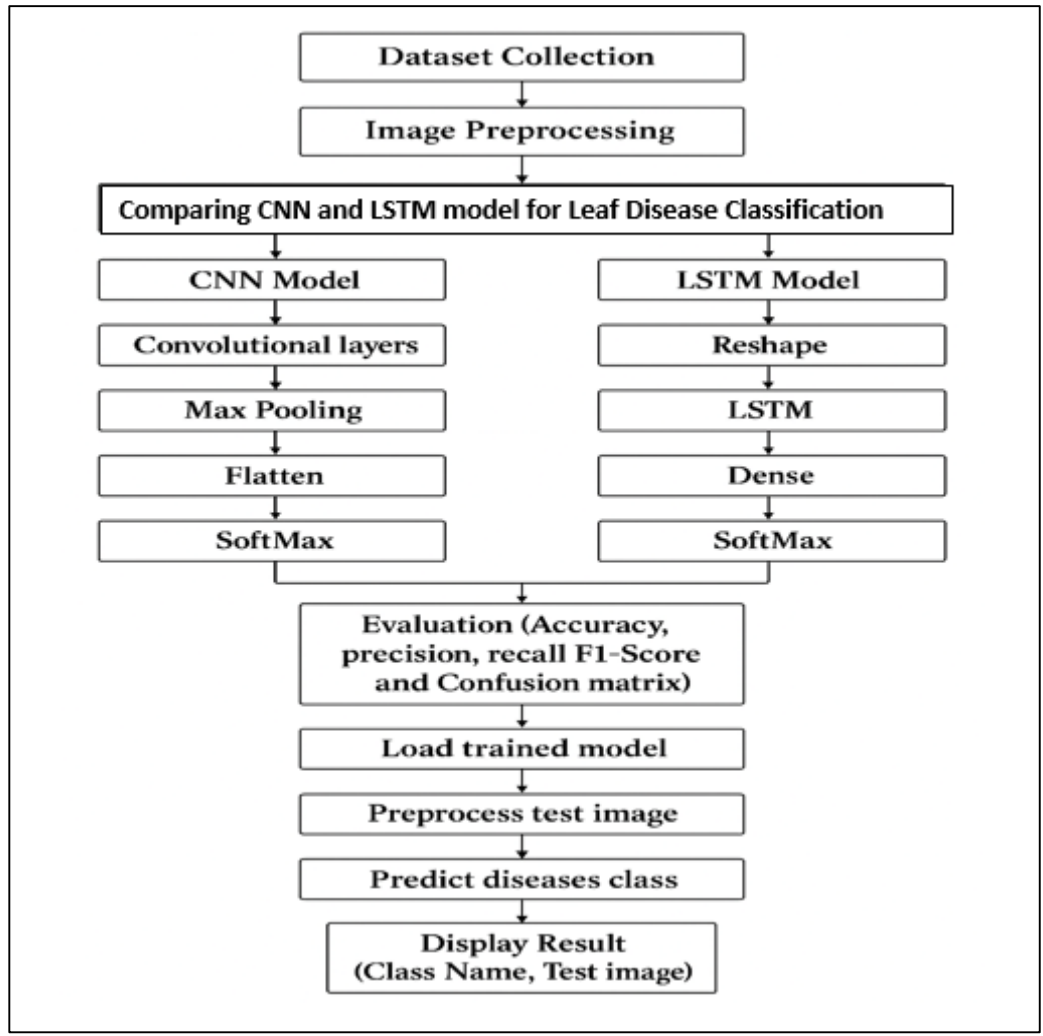

**Figure 2.** Proposed Model Architecture for Multi-Leaf Disease Classification

## 3.3 Algorithms

### 3.3.1 Convolutional Neural Network (CNN)

The proposed convolutional neural network (CNN) model is designed to perform multi-class image classification on RGB input images of size 128×128×3. The architecture consists of a series of convolutional and max-pooling layers arranged in blocks, followed by fully connected layers for final classification. Each convolutional block contains two convolutional layers with ReLU activation and a subsequent max-pooling layer. The number of filters increases progressively from 32 to 512 across the blocks, allowing the network to learn hierarchical features from low-level textures to high-level abstractions. The first convolutional block includes two Conv2D layers with 32 filters each, followed by a max-pooling layer. This is succeeded by similar blocks with 64, 128, 256, and 512 filters, each employing 3×3 kernels and a stride of 1. Max-pooling layers with a pool size of 2×2 are used to downsample feature maps and reduce computational complexity. After the final convolutional block, the feature maps are flattened and passed through a dense layer of 1500 neurons with ReLU activation. Dropout regularisation is applied before and after this dense layer to mitigate overfitting. The output layer is a dense layer with 38 neurons and a SoftMax activation function, suitable for multi-class classification. The total number of trainable parameters in the model is 7,842,762. The model summary is depicted in Table I.

**Table 1.** CNN Model Summary

| Layer (Type) | Output Shape | Number of Parameters | Description |
|---|---|---|---|
| Conv2d | (128, 128, 32) | 896 | Applies 32 filters of size 3×3; extracts low-level features from input image. |
| conv2d_1 | (128, 128, 32) | 9,248 | Another 32 filters increase feature abstraction at the initial resolution. |
| max_pooling2d | (63, 63, 32) | 0 | Reduces spatial size to control overfitting and computational load. |
| conv2d_2 | (63, 63, 64) | 18,496 | Doubles filter count to 64; detects more complex patterns. |
| conv2d_3 | (61, 61, 64) | 36,928 | Reinforces learned features with another 64 filters. |
| max_pooling2d_1 | (30, 30, 64) | 0 | Further down, sampling of feature maps. |
| conv2d_4 | (30, 30, 128) | 73,856 | Increases depth to 128; learns mid-level patterns like corners and textures. |
| conv2d_5 | (28, 28, 128) | 147,584 | Reinforces patterns using the same number of filters. |
| max_pooling2d_2 | (14, 14, 128) | 0 | Reduces height and width by half. |
| conv2d_6 | (14, 14, 256) | 295,168 | Deepens feature representation to 256 filters. |

| conv2d_7 | (12, 12, 256) | 590,080 | Second 256-filter convolution in this block for deeper abstraction. |
|---|---|---|---|
| max_pooling2d_3 | (6, 6, 256) | 0 | Downsampled again to preserve key spatial features. |
| conv2d_8 | (6, 6, 512) | 1,180,160 | Sharp increase to 512 filters; focuses on complex, class-specific features. |
| conv2d_9 | (4, 4, 512) | 2,359,808 | Deepens pattern abstraction at full filter depth |
| max_pooling2d_4 | (2, 2, 512) | 0 | Final pooling layer before fully connected layers. |
| dropout | (2, 2, 512) | 0 | Prevents overfitting by randomly zeroing some neurons during training. |
| flatten | (2048) | 0 | Flattens the 3D feature map to 1D vector for dense layers. |
| dense | (1500) | 3,073,500 | Fully connected layer with 1500 neurons for feature interpretation. |
| dropout_1 | (1500) | 0 | Additional dropout regularisation to avoid overfitting |
| dense_1 | (38) | 57,038 | Final output layer with 38 neurons; uses softmax activation for classification. |
| Total | | 7,842,762 | |

- Total Parameters: 7,842,762 (29,92 MB)

- Trainable Parameters: 7,842,762 (29,92 MB)

- Non-trainable parameters: 0 (0.00 B)

- Output Classes: 38 (different crop disease categories)

### 3.3.2 Long Short-Term Memory (LSTM)

In addition to convolutional architectures, a recurrent neural network (RNN) model based on Long Short-Term Memory (LSTM) was developed for sequence-based classification tasks. The model is designed to process time-dependent sequential data patterns, such as temporal features derived from crop health metrics The architecture consists of a single LSTM layer with 128 memory units, followed by two fully connected layers. The LSTM layer captures temporal dependencies and retains long-term contextual information within the input sequence. It is followed by a dense layer with 128 neurons using ReLU activation to enhance non-linearity and feature representation. The final dense layer employs a softmax activation function and consists of 38 output neurons corresponding to the number of target classes. The model contains a total of 742,822 trainable parameters and no non-trainable parameters. Table 2 summarises model summary of LSTM.

**Table 2.** LSTM Model Summary

| Layer (Type) | Output Shape | Number of Parameters | Description |
|---|---|---|---|
| LSTM (lstm_19) | (None, 128) | 721,408 | Long Short-Term Memory layer with 128 units; captures temporal dependencies. |
| Dense (dense_22) | (None, 128) | 16,512 | Fully connected layer with 128 units for nonlinear transformation. |
| Dense (dense_23) | (None, 38) | 4,902 | Final classification layer with 38 output classes using softmax activation. |
| Total | | 742,822 | |

- Total Trainable Parameters: 742,822 (2.83 MB)

- Trainable parameters: 742,822 (2.83 MB)

- Non-trainable parameters: (0.00 B)

## 4. Results and Discussion

### 4.1. Performance Evaluation

Agriculture plays an important role in ensuring global food security, making the timely and accurate detection of plant diseases essential for maintaining crop quality and yield. This study explores a range of machine learning (ML) and deep learning (DL) approaches used to detect and classify plant diseases based on symptoms visible on leaves and crops. By utilizing automated classification techniques, farmers can accurately identify diseases, reducing the reliance on manual observation and allowing for earlier and more effective interventions. Traditionally, disease identification has required specialised knowledge in plant pathology, with manual inspections often being slow and susceptible to errors. In contrast, ML and DL methods offer the potential for accurate, large-scale, and automated disease recognition. Establishing an early warning system for plant diseases could greatly benefit agriculture by minimising crop loss and improving the efficiency of treatment strategies. The field of automated plant disease detection is advancing quickly, utilising image analysis and classification methods to monitor plant health across wide areas. Factors such as the size of datasets, the number of disease categories, image preprocessing techniques, classification methods, and evaluation metrics all influence the performance of these systems.

### Confusion Matrix Analysis

The confusion matrix with a detailed summary of how well the model performs is shown below in Figure 3 and 5.

### 4.2.1 CNN Performance Evaluation

- **CNN Confusion Matrix**

The convolutional neural network (CNN) model, which is well known for its ability to learn spatial features from images, attains high accuracy in classification. The confusion matrix of CNN presents good performance, especially in distinguishing specific disease classes. However, minor misclassifications occur in visually similar disease types, suggesting the need for further fine-tuning and augmentation methods. Figure 4 shows the CNN classification report.

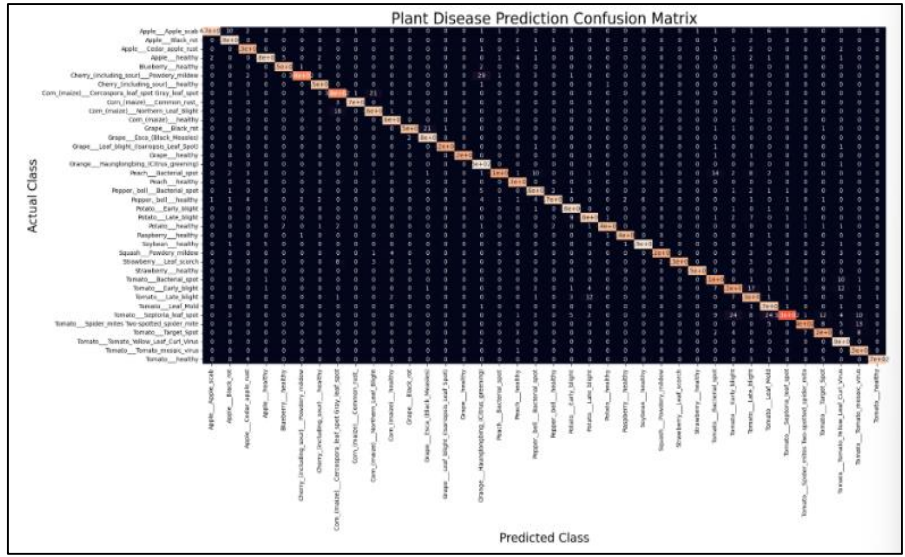

**Figure 3.** CNN Confusion Matrix

## CNN Classification Report

|  | precision | recall | f1-score | support |
|---|---|---|---|---|
| Apple___Apple_scab | 0.99 | 0.93 | 0.96 | 504 |
| Apple___Black_rot | 0.96 | 0.98 | 0.97 | 497 |
| Apple___Cedar_apple_rust | 0.96 | 0.98 | 0.97 | 440 |
| Apple___healthy | 0.97 | 0.96 | 0.96 | 502 |
| Blueberry___healthy | 0.98 | 0.98 | 0.98 | 454 |
| Cherry_(including_sour)___Powdery_mildew | 0.99 | 0.90 | 0.94 | 421 |
| Cherry_(including_sour)___healthy | 0.99 | 1.00 | 0.99 | 456 |
| Corn_(maize)___Cercospora_leaf_spot Gray_leaf_spot | 0.92 | 0.94 | 0.93 | 410 |
| Corn_(maize)___Common_rust_ | 1.00 | 0.99 | 0.99 | 477 |
| Corn_(maize)___Northern_Leaf_Blight | 0.95 | 0.96 | 0.95 | 477 |
| Corn_(maize)___healthy | 0.99 | 1.00 | 0.99 | 465 |
| Grape___Black_rot | 0.99 | 0.95 | 0.97 | 472 |
| Grape___Esca_(Black_Measles) | 0.95 | 1.00 | 0.97 | 480 |
| Grape___Leaf_blight_(Isariopsis_Leaf_Spot) | 1.00 | 0.99 | 0.99 | 430 |
| Grape___healthy | 1.00 | 1.00 | 1.00 | 423 |
| Orange___Haunglongbing_(Citrus_greening) | 0.91 | 1.00 | 0.95 | 503 |
| Peach___Bacterial_spot | 0.99 | 0.90 | 0.94 | 459 |
| Peach___healthy | 0.95 | 1.00 | 0.98 | 432 |
| Pepper,_bell___Bacterial_spot | 0.94 | 0.97 | 0.96 | 478 |
| Pepper,_bell___healthy | 0.99 | 0.94 | 0.96 | 497 |
| Potato___Early_blight | 0.95 | 0.99 | 0.97 | 485 |
| Potato___Late_blight | 0.96 | 0.95 | 0.96 | 485 |
| Potato___healthy | 0.99 | 0.97 | 0.98 | 456 |
| Raspberry___healthy | 0.97 | 0.99 | 0.98 | 445 |
| Soybean___healthy | 0.99 | 0.98 | 0.98 | 505 |
| Squash___Powdery_mildew | 0.99 | 0.97 | 0.98 | 434 |
| Strawberry___Leaf_scorch | 1.00 | 0.96 | 0.98 | 444 |
| Strawberry___healthy | 1.00 | 0.98 | 0.99 | 456 |
| Tomato___Bacterial_spot | 0.92 | 0.97 | 0.94 | 425 |
| Tomato___Early_blight | 0.91 | 0.88 | 0.89 | 480 |
| Tomato___Late_blight | 0.86 | 0.94 | 0.90 | 463 |
| Tomato___Leaf_Mold | 0.91 | 0.99 | 0.95 | 470 |
| Tomato___Septoria_leaf_spot | 0.98 | 0.76 | 0.85 | 436 |
| Tomato___Spider_mites Two-spotted_spider_mite | 0.97 | 0.93 | 0.95 | 435 |
| Tomato___Target_Spot | 0.92 | 0.92 | 0.92 | 457 |
| Tomato___Tomato_Yellow_Leaf_Curl_Virus | 0.91 | 1.00 | 0.95 | 490 |
| Tomato___Tomato_mosaic_virus | 0.93 | 1.00 | 0.96 | 448 |
| Tomato___healthy | 0.98 | 0.97 | 0.98 | 481 |
|  |  |  |  |  |
| accuracy |  |  | 0.96 | 17572 |
| macro avg | 0.96 | 0.96 | 0.96 | 17572 |
| weighted avg | 0.96 | 0.96 | 0.96 | 17572 |

**Figure 4.** CNN Classification Report

## 4.2.2 LSTM Performance Evaluation

- **LSTM Confusion Matrix**

The LSTM model, which is typically used for sequential data processing, is modified here for image-based classification by taking advantage of extracted feature representations. The confusion matrix for LSTM indicates that while it captures patterns effectively, it exhibits minor performance degradation compared to CNN, particularly in closely related disease classes. The LSTM classification report is shown in Figure 6.

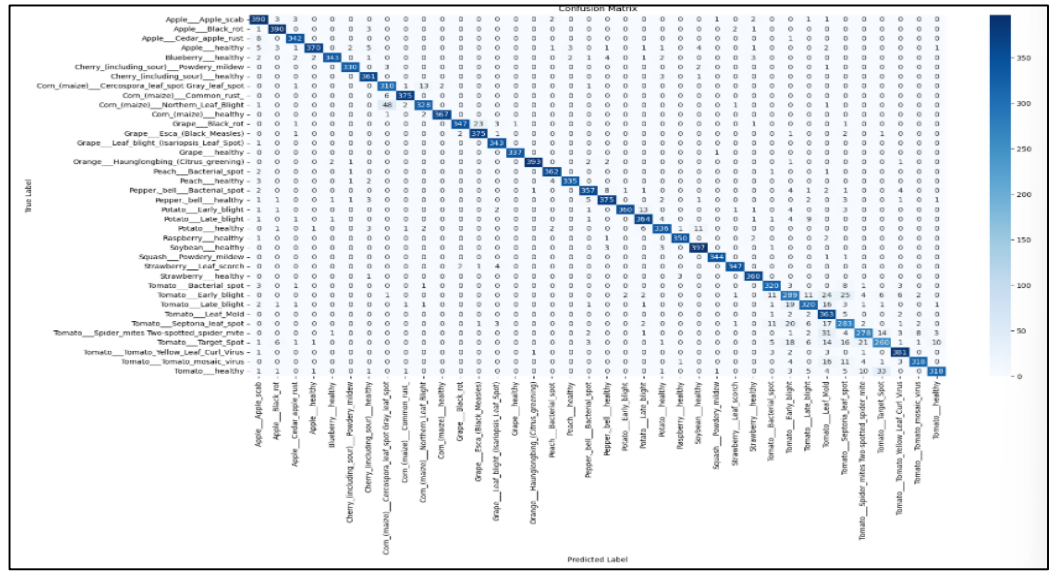

**Figure 5.** LSTM Confusion Matrix

- **LSTM Classification Report**

| | precision | recall | f1-score | suppor |
|---|---|---|---|---|
| t | | | | |
| Apple___Apple_scab | 0.91 | 0.97 | 0.94 | 403 |
| Apple___Black_rot | 0.96 | 0.98 | 0.97 | 397 |
| Apple___Cedar_apple_rust | 0.96 | 0.97 | 0.97 | 351 |
| Apple___healthy | 0.99 | 0.92 | 0.95 | 401 |
| Blueberry___healthy | 0.99 | 0.94 | 0.96 | 363 |
| Cherry_(including_sour)___Powdery_mildew | 0.98 | 0.98 | 0.98 | 336 |
| Cherry_(including_sour)___healthy | 0.95 | 0.99 | 0.97 | 365 |
| Corn_(maize)___Cercospora_leaf_spot Gray_leaf_spot | 0.84 | 0.95 | 0.89 | 328 |
| Corn_(maize)___Common_rust_ | 0.98 | 0.98 | 0.98 | 381 |
| Corn_(maize)___Northern_Leaf_Blight | 0.95 | 0.86 | 0.90 | 381 |
| Corn_(maize)___healthy | 0.99 | 0.99 | 0.99 | 370 |
| Grape___Black_rot | 0.99 | 0.92 | 0.95 | 377 |
| Grape___Esca_(Black_Measles) | 0.94 | 0.98 | 0.96 | 383 |
| Grape___Leaf_blight_(Isariopsis_Leaf_Spot) | 0.96 | 1.00 | 0.98 | 344 |
| Grape___healthy | 1.00 | 1.00 | 1.00 | 338 |
| Orange___Haunglongbing_(Citrus_greening) | 0.99 | 0.98 | 0.99 | 402 |
| Peach___Bacterial_spot | 0.98 | 0.99 | 0.98 | 367 |
| Peach___healthy | 0.99 | 0.97 | 0.98 | 345 |
| Pepper,_bell___Bacterial_spot | 0.95 | 0.93 | 0.94 | 382 |
| Pepper,_bell___healthy | 0.95 | 0.94 | 0.95 | 397 |
| Potato___Early_blight | 0.99 | 0.93 | 0.96 | 387 |
| Potato___Late_blight | 0.93 | 0.94 | 0.93 | 387 |
| Potato___healthy | 0.95 | 0.92 | 0.94 | 364 |
| Raspberry___healthy | 0.99 | 0.98 | 0.98 | 356 |
| Soybean___healthy | 0.95 | 0.98 | 0.97 | 404 |
| Squash___Powdery_mildew | 0.99 | 0.99 | 0.99 | 346 |
| Strawberry___Leaf_scorch | 0.98 | 0.98 | 0.98 | 354 |
| Strawberry___healthy | 0.97 | 0.99 | 0.98 | 364 |
| Tomato___Bacterial_spot | 0.90 | 0.94 | 0.92 | 340 |
| Tomato___Early_blight | 0.77 | 0.75 | 0.76 | 384 |
| Tomato___Late_blight | 0.88 | 0.86 | 0.87 | 370 |
| Tomato___Leaf_Mold | 0.73 | 0.97 | 0.83 | 375 |
| Tomato___Septoria_leaf_spot | 0.76 | 0.81 | 0.79 | 349 |
| Tomato___Spider_mites Two-spotted_spider_mite | 0.86 | 0.80 | 0.83 | 348 |
| Tomato___Target_Spot | 0.82 | 0.71 | 0.76 | 365 |
| Tomato___Tomato_Yellow_Leaf_Curl_Virus | 0.94 | 0.97 | 0.95 | 392 |
| Tomato___Tomato_mosaic_virus | 0.96 | 0.89 | 0.92 | 358 |
| Tomato___healthy | 0.95 | 0.83 | 0.88 | 385 |
| accuracy | | | 0.93 | 14039 |
| macro avg | 0.94 | 0.93 | 0.93 | 14039 |
| weighted avg | 0.94 | 0.93 | 0.93 | 14039 |

**Figure 6.** LSTM Classification Report

## 4.2.3 Performance Comparison of CNN and LSTM

The evaluation metrics are used to compare the performance of the Convolutional Neural Network (CNN) and Long Short-Term Memory (LSTM) models. These metrics include Accuracy, Precision, Recall, and F1-Score, which are standard in multi-class classification tasks.

**Table 3.** Performance Comparison of CNN and LSTM

| | Model | Accuracy | Precision | Recall | F1-Score |
|---|---|---|---|---|---|
| 1 | CNN | 0.9908 | 0.9913 | 0.9913 | 0.9913 |

| 2 | LSTM | 0.9338 | 0.9300 | 0.9300 | 0.9300 |
|---|------|--------|--------|--------|--------|

The CNN model achieved an accuracy of 99.08%, indicating a high level of correct predictions. Precision, recall, and F1-score are all measured at 0.9913, demonstrating consistent performance in identifying the correct class labels with minimal false positives and false negatives. In contrast, the LSTM model recorded an accuracy of 93.38%, with precision, recall, and F1-score each at 0.9300. Although still effective, the LSTM model exhibited a relatively lower classification performance compared to the CNN as shown in Table 3.

**Evaluation Metrics**

The classification performance was evaluated using the following formulas, based on the confusion matrix elements:

- **True Positives (TP):** Correctly predicted positive cases.

- **False Positives (FP):** Incorrectly predicted positive cases.

- **True Negatives (TN):** Correctly predicted negative cases.

- **False Negatives (FN):** Incorrectly predicted negative cases.

The formulas used are:

1. **Accuracy**

$$Accuracy = \frac{TP+TN}{TP+TN+FP+FN}$$

It measures the overall correctness of the model. CNN has the highest accuracy (99.07%), while LSTM performs the worst (93.38%).

2. **Precision**

$$Precision = \frac{TP}{TP+FP}$$

Precision indicates how many of the predicted positive cases were correct. CNN again performs best, suggesting it makes fewer false positive errors.

3. **Recall**

$$Recall = \frac{TP}{TP + FN}$$

Recall measures how well the model identifies actual positive cases. Higher recall means fewer missed classifications. CNN have high recall, indicating good detection capability.

4. **F1-Score**

$$F1\ Score = 2 * \frac{Precision * Recall}{Precision + Recall}$$

The F1-score balances precision and recall. CNN has the best F1-score (0.9913), confirming its strong classification performance, while LSTM has the lowest (0.9300), indicating poor predictive ability.

**CNN Model:** Loss and Accuracy

The CNN model's training loss decreases steadily over the epochs, indicating effective learning. However, the validation loss fluctuates significantly, suggesting potential overfitting. The accuracy plots shows that while training accuracy improves consistently, validation accuracy exhibits variations, implying that the model may not generalise well to unseen data. This could be due to limited dataset size, insufficient regularisation, or an imbalance in training and validation data. Figure 7 illustrates the Loss and the Accuracy of the CNN.

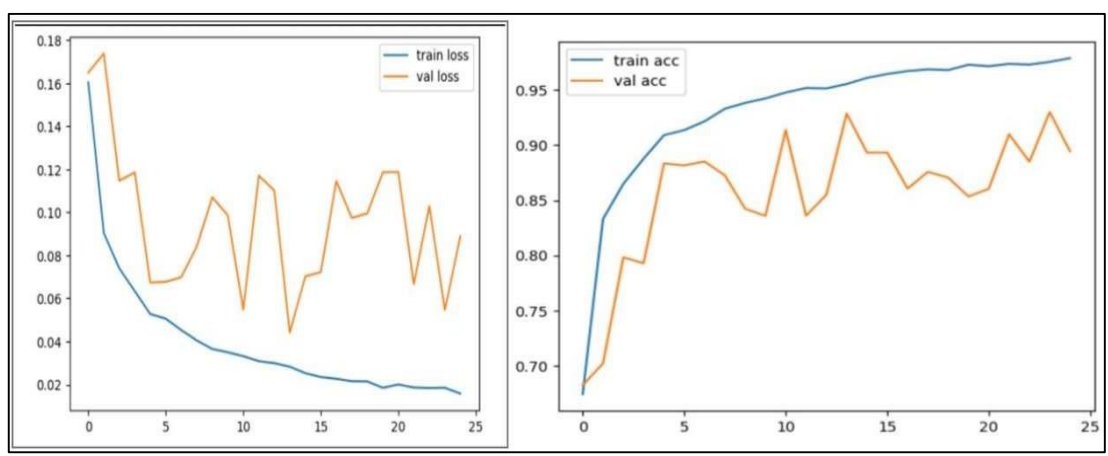

**Figure 7.** CNN Model: Loss and Accuracy

**LSTM Model:** Loss and Accuracy

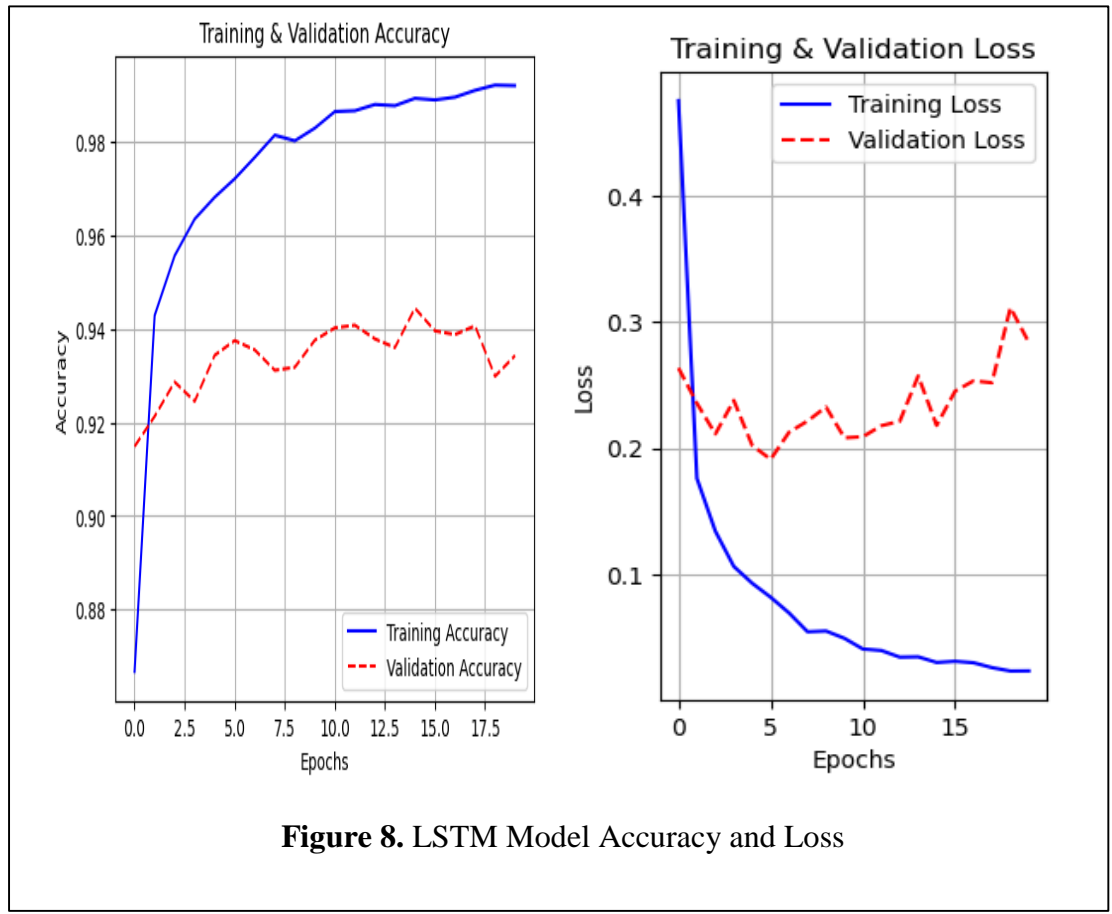

**Figure 8.** LSTM Model Accuracy and Loss

For the LSTM model, both training and validation accuracy improve gradually, demonstrating stable learning. The loss curve shows a smooth decline, indicating that the model is effectively minimizing errors. Unlike the CNN model, the LSTM exhibits less fluctuation in validation performance, suggesting better generalization. This is expected since LSTMs are well-suited for sequential data and may better capture temporal dependencies in the dataset. Figure 8 illustrates the Accuracy and Loss of the LSTM.

**4.3 Disease Classification Results**

The image shows the result of a disease classification system applied to a Potato leaf. On the left side, Figure 9 presents the input test image, which displays a Potato leaf with visible signs of deterioration, including browning, yellowing, and damaged tissue. On the right side, Figure 10 shows the predicted disease classification result for the same image. The system has identified the disease as Potato Early Blight. This condition, caused by the fungus Alternaria solani, commonly affects older leaves and is characterised by dark, concentric spots and yellowing around the affected areas. The prediction aligns with the symptoms seen in the test image, confirming the presence of Early Blight on the Potato leaf.

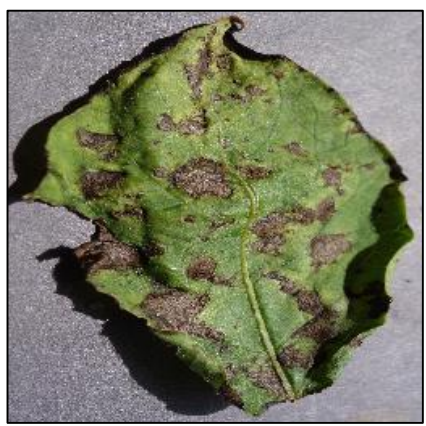 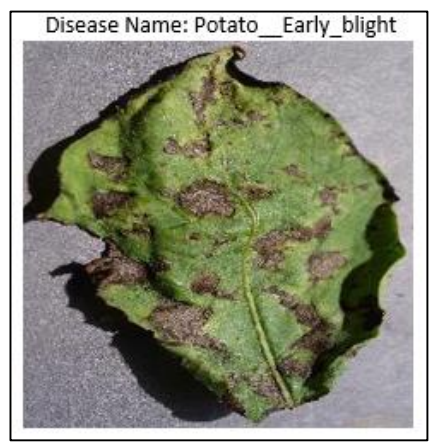

**Figure 9.** Input Test Image          **Figure 10.** Predicted Disease Classification

## 4.4 Discussion

The results obtained confirms that the plant disease classification learning model using CNN outperforms the traditional machine learning techniques.

### Disease vs. Healthy Leaf Classification

The model successfully differentiates between healthy and diseased leaves, as shown in the results. The test image was correctly classified as Potato Early blight, highlighting the model's robustness in identifying plant diseases. The deep learning approach significantly reduces false positives and negatives compared to conventional classifiers, ensuring reliable disease detection.

## 5.   Conclusion

In conclusion, the results clearly show that Convolutional Neural Networks (CNNs) are more effective than Long Short-Term Memory (LSTM) models for classifying plant diseases based on leaf images. The CNN model achieved outstanding performance, with an accuracy of 0.9908, a precision and recall of 0.9913, and an F1-score of 0.9913. This indicates that CNNs are well-suited for tasks like identifying diseases such as Early Blight, as they are excellent at recognising visual patterns such as texture, shape, and colour, which are essential for diagnosing plant diseases from leaf images. The LSTM model, which is typically used for sequential or time-series data, showed weaker results, with an accuracy of 0.9338 and lower precision, recall, and F1-scores around 0.9300. LSTMs, while good at understanding sequences, do not perform as well for image-based tasks where recognising spatial relationships is essential. This confirms that CNNs are the preferred choice for image classification tasks like plant disease detection, offering a more reliable and accurate solution compared to LSTMs.

**Author's biography**

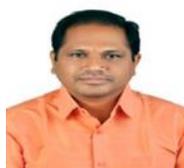

**Dr. Srinivas Kanakala**, working as Senior Assistant Professor in the Department of Computer Science and Engineering, VNR Vignana Jyothi Institute of Engineering & Technology, Hyderabad, India. He has completed his B. Tech and M.Tech in Computer Science and Engineering from the Jawaharlal Nehru Technological University Hyderabad. He has completed Ph.D from Osmania University , Hyderabad in the year 2018. He has 20 years of academic and research experience. His major field of study is Machine Learning, Mobile Computing, Data Science. He has guided many Undergraduate and Post-graduate students. He has around 40 publications indexed in Scopus and web of science, in various international journals and conferences. He is a Life member of ISTE and CSI and working as a reviewer for various journals and conferences.

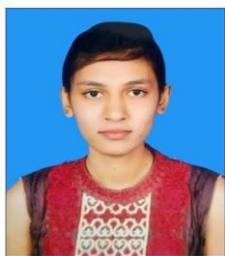

**Ningappa Sneha** pursuing M.Tech in Software Engineering at Vallurupalli Nageswara Rao Vignana Jyothi Institute of Engineering and Technology, Hyderabad. Completed B.Tech in Computer Science and Engineering from DRK College of Engineering and Technology. And also completed a Java Full Stack Development course at J-Spiders Academy, to further enhance technical skill set. Research interests include Deep Learning, Computer Vision, and Agricultural Informatics. Worked on impactful academic projects such as "Detection and Classification of Tomato Leaf Diseases" using ensemble models of LSTM and CNN, and "Detection of Diabetic Retinopathy Using Deep CNN with Data Augmentation on Cropped Fundus Images." Both projects showcase a commitment towards practical, AI-driven healthcare and agriculture solutions. And also presented research work at a conference and is currently awaiting publication of the research.